\documentclass[preprint]{article}

\usepackage{neurips_2024}

\usepackage{algorithm}
\usepackage{algorithmic}
\usepackage[utf8]{inputenc} 
\usepackage[T1]{fontenc}    
\usepackage{url}            
\usepackage{booktabs}       
\usepackage{amsfonts}       
\usepackage{nicefrac}       
\usepackage{microtype}      
\usepackage{xcolor}         
\definecolor{brickred}{rgb}{0.8, 0.25, 0.33}
\definecolor{bostonuniversityred}{rgb}{0.8, 0.0, 0.0}
\definecolor{brightmaroon}{rgb}{0.76, 0.13, 0.28}
\definecolor{candyapplered}{rgb}{1.0, 0.03, 0.0}
\definecolor{carminered}{rgb}{1.0, 0.0, 0.22}
\definecolor{coralred}{rgb}{1.0, 0.25, 0.25}
\definecolor{cornellred}{rgb}{0.7, 0.11, 0.11}
\definecolor{citecolor}{RGB}{17,80,197}
\usepackage[colorlinks,     
            linkcolor=carminered,
            anchorcolor=blue,
            citecolor=citecolor,   
            ]{hyperref}
\usepackage{amsmath}
\usepackage{amssymb}
\usepackage{mathtools}
\usepackage{amsthm}
\usepackage{multirow}
\usepackage{tabularx}
\usepackage{microtype}
\usepackage{graphicx}
\usepackage{subfigure}
\usepackage{booktabs} 
\usepackage{pifont}
\usepackage{empheq}
\usepackage{amssymb}
\usepackage{pifont}%
\usepackage{colortbl}

\newtheorem{theorem}{Theorem}[section]

\newtheorem{remark}[theorem]{Remark}

\DeclareMathOperator*{\argmin}{arg\,min}

\newlength\savewidth

\definecolor{codeblue}{rgb}{0.25, 0.5, 0.5}
\definecolor{codekw}{rgb}{0.35, 0.35, 0.75}
\definecolor{Gray}{gray}{0.95}

\newcommand{\gr}{\rowcolor[gray]{.95}}

\title{
A Convex-optimization-based Layer-wise Post-training Pruner for Large Language Models
}

\author{%
  Pengxiang Zhao\thanks{Equal contribution} \\
  Department of Mathematics\\
  The University of Hong Kong\\
  \texttt{pengxiangzhao@connect.hku.hk} \\
  \And
  Hanyu Hu$^*$ \\
  Department of Mathematics \\
  The University of Hong Kong \\
  \texttt{hhy1224@connect.hku.hk} \\
  \AND
  Ping Li \\
  System AI Innovation Lab \\
  Huawei Cloud \\
  \texttt{liping61@huawei.com} \\
  \And
  Yi Zheng \\
  System AI Innovation Lab \\
  Huawei Cloud \\
  \texttt{zhengyi29@huawei.com} \\
  \AND
  Zhefeng Wang \\
  System AI Innovation Lab \\
  Huawei Cloud \\
  \texttt{wangzhefeng@huawei.com} \\
  \And
  Xiaoming Yuan\thanks{Corresponding author} \\
  Department of Mathematics \\
  The University of Hong Kong \\
  \texttt{xmyuan@hku.hk} \\
}

\begin{document}

\maketitle

\begin{abstract}
  Pruning is a critical strategy for compressing trained large language models (LLMs), aiming at substantial memory conservation and computational acceleration without compromising performance. However, existing pruning methods often necessitate inefficient retraining for billion-scale LLMs or rely on heuristic methods such as the optimal brain surgeon framework, which degrade performance. 
  In this paper, we introduce FISTAPruner, the first post-training pruner based on convex optimization models and algorithms. Specifically, we propose a convex optimization model incorporating $\ell_1$ norm to induce sparsity and utilize the FISTA solver for optimization. FISTAPruner incorporates an intra-layer cumulative error correction mechanism and supports parallel pruning. We comprehensively evaluate FISTAPruner on models such as OPT, LLaMA, LLaMA-2, and LLaMA-3 with 125M to 70B parameters under unstructured and 2:4 semi-structured sparsity, demonstrating superior performance over existing state-of-the-art methods across various language benchmarks.
\end{abstract}

\section{Introduction}
In recent years, large language models (LLMs) have revolutionized natural language processing fields, achieving impressive results in tasks such as machine translation, sentiment analysis, question answering, and text generation \citep{lyu2023new, yao2023empowering, zhang2023prompting, zhang2023sentiment, wang2023augmenting, arefeen2024leancontext, li2024pre}. Advanced LLMs such as OpenAI's GPT-4~\citep{openai2023gpt}, Meta's LLaMA-3~\citep{meta2023llama3}, and Google's Gemini \citep{team2023gemini} excel in generating coherent text with extensive parameters. However, the growth in model sizes outpaces hardware improvements, posing significant deployment and inference challenges \citep{steiner2023model}. For example, operating OPT-175B \citep{zhang2022opt} requires over 320GB of memory and at least five 80GB A100 GPUs for loading its parameters in FP16 precision. This challenge becomes more pronounced in environments with limited resources, such as mobile devices, edge computing systems, and real-time applications. Consequently, there has been considerable interest in compressing LLMs to enhance their efficiency and practicality for deployment across various applications.

Pruning is a key method for compressing LLMs, aiming to eliminate redundant weights to reduce model size and computational demands while striving to maintain performance. Methods such as those in \citep{huang2020convolution, ma2023llm_pruner, zhang2023lottery} require a retraining phase post-pruning, which is inefficient for billion-scale LLMs. Recent developments, including SparseGPT~\citep{frantar2023sparsegpt} and Wanda \citep{sun2023simple}, employ one-shot post-training pruning techniques for LLMs. These methods, however, rely on the heuristic-based optimal brain surgeon (OBS) framework~\citep{hassibi1992second} or utilize heuristic-based pruning metric to determine which weights to prune, potentially compromising performance.

\begin{figure}
    \centering
    \includegraphics[width=\linewidth]{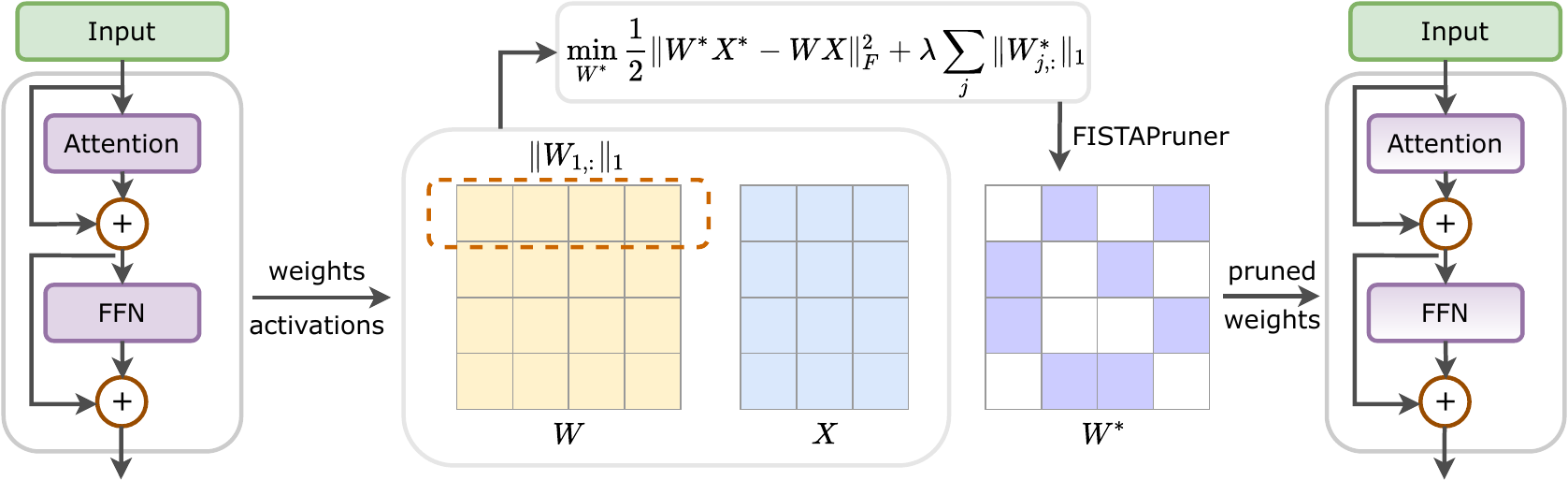}
    \caption{Overview of the proposed FISTAPruner. Given a weight matrix $W$ and its corresponding input feature activation $X$, we employ the proposed convex optimization model, utilizing FISTA, to derive the pruned weights.}
    \label{fig:overview}
\end{figure}

In this work, we introduce a novel convex optimization model for layer-wise post-training pruning of LLMs. Figure \ref{fig:overview} provides an overview of our method, which is applied to each linear operator, such as \(W_K\), \(W_Q\), \(W_V\), and $W_O$ within the Transformer's attention blocks \citep{vaswani2017attention}. We employ the Frobenius norm of the difference between the outputs obtained from the dense and pruned weights to quantify the output error. Additionally, we integrate an $\ell_1$-norm regularization term, the optimal convex approximation of the $\ell_0$-norm~\citep{candes2006robust}, into each row of weights to promote sparsity. The solutions of the proposed optimization model demonstrate a balanced trade-off between output error and sparsity, governed by our proposed adaptive tuning method that meticulously adjusts the hyperparameter $\lambda$. To solve this optimization problem efficiently, we utilize the Fast Iterative Shrinkage-Thresholding Algorithm (FISTA) \citep{beck2009fast}, which ensures a convergence rate of $O(1/k^2)$. Following this, we name our proposed method FISTAPruner.
 
In addition, our approach effectively addresses the cumulative error resulting from compression within the decoder layers by incorporating an error correction mechanism. Specifically, there exists a discrepancy between the outputs of the dense and the pruned weights. This error accumulates since the output from one pruned weight serves as the input for the subsequent operator. To mitigate this accumulated error, FISTAPruner sequentially prunes the weights of each linear operator within a decoder layer. It utilizes the output from the pruned weights of the preceding operator as the input activation for the ongoing pruning process, thereby minimizing the gap between the output weight being pruned and its counterpart in the dense model. 
Moreover, FISTAPruner treats each decoder layer as an independent pruning unit, supporting the simultaneous pruning of multiple decoder layers, which significantly enhances efficiency.

We empirically evaluate FISTAPruner on the widely adopted OPT \citep{zhang2022opt}, LLaMA~\citep{touvron2023llama}, and LLaMA-2~\citep{touvron2023llama2} model families, as well as the latest LLaMA-3 \citep{touvron2023llama} models. FISTAPruner’s layer-by-layer pruning implementation allows for the pruning of these LLMs ranging from 125M to 70B parameters on a single NVIDIA A100 GPU with 40GB of memory. Our results confirm that FISTAPruner can efficiently create sparse networks from pretrained LLMs without retraining. Moreover, our approach exceeds the performance of state-of-the-art methods such as SparseGPT and Wanda across various language benchmarks. We believe our work sets a new direction and baseline for future research in this area and encourages further exploration into understanding sparsity in LLMs with the tools of convex optimization.

\section{Background and Related Work}

\textbf{Pruning of LLMs.}
Pruning is a widely used strategy to compress LLMs by generating sparse weight matrices under unstructured, semi-structured, and structured sparsity based on calibration data. 
Unstructured sparsity of rate $s\%$, eliminates $s\%$ of the entries in a weight matrix. Semi-structured sparsity with proportion $n:m$ maintains a fixed overall sparsity level $n/m$, and allows at most $n$ non-zero entries in every group of $m$ consecutive entries. 
Pruning weights into semi-structured sparsity, especially with proportion 2:4, could yield up to 2$\times$ inference speedup using NVIDIA GPUs with the Ampere architecture \citep{mishra2021accelerating} and hence is of particular interest.
Structured sparsity, which zeroes entire rows or columns, offers significant computational and memory benefits but can lead to greater performance losses. Depending on whether to incorporate another round of training after pruning to recover performance, pruning schemes could be further classified into pruning with retraining and one-shot pruning.

\textbf{Pruning with Retraining.}
Traditional pruning pipelines often include a retraining step to offset performance losses \citep{huang2020convolution, ma2023llm_pruner, zhang2023lottery}. However, the sheer scale of LLMs makes this additional retraining costly in terms of both time and computational resources. \citep{dinh2020sparsity, holmes2021nxmtransformer, xie2023hollownerf} integrate retraining directly into the pruning process by targeting the minimization of the highly non-convex loss function related to the calibration dataset, using the alternating direction method of multipliers (ADMM) to derive pruned weights. 
Nonetheless, this approach imposes significant computational demands and the use of ADMM in non-convex optimization often results in unstable performance \citep{he20121}.

\textbf{One-Shot Pruning.}
One-shot pruning offers a straightforward alternative, eliminating the need for post-pruning retraining. These methods prune LLMs in a single step, simplifying implementation and reducing both time and computational demands. Consequently, various one-shot pruning algorithms have been developed under different sparsity frameworks. For structured pruning, SliceGPT~\citep{ashkboos2024slicegpt} and Eigenpruning \citep{vergara2024eigenpruning} utilize singular value decompositions to prune singular values of weight matrices and reduce model dimensions. ZipLM \citep{kurtic2024ziplm} adopts an OBS-based approach for structured pruning and updates remaining weights to maintain performance.
Our proposed FISTAPruner focuses on unstructured and semi-structured pruning, and thus is orthogonal to these structured pruning methods, enabling further model compression.
For unstructured and semi-structured pruning, SparseGPT \citep{frantar2023sparsegpt} and ISC \citep{shao2024one} leverage the OBS framework to calculate saliency for each entry using the inverse Hessian of the loss metric, based on which pruning masks are generated and weights updated. Wanda \citep{sun2023simple} implements a heuristic approach, removing weights based on the product of their magnitudes and activations without compensation. \citep{bovza2024fast} employs ADMM to optimize weight updates under iteratively refined pruning masks chosen through heuristic methods. 
These strategies adopt a layer-wise pruning strategy, where errors between the pruned output and the original output of each operator accumulates.
Moreover, due to their heuristic nature, the performances of the pruned models are unstable and compromised.
In contrast, our proposed FISTAPruner introduces a novel layer-wise one-shot pruning approach by formulating it as a convex optimization problem with intra-layer error corrections. It employs the FISTA solver to efficiently compute optimal pruned weights, providing theoretical guarantees for performance stability and effectiveness.
 
\section{Methodology}
In this section, we present our proposed post-training pruning method, FISTAPruner, which comprises three main components. First, we identify the error accumulation problem in layer-wise pruning and propose an intra-layer error correction mechanism to address it. Based on this mechanism, we establish a novel convex optimization model tailored for layer-wise pruning. We then detail the steps for solving this optimization problem using the FISTA solver. Finally, we describe our adaptive method that finely tunes the hyperparameter $\lambda$ in our model, aiming to minimize the output error between dense and pruned operators while achieving the desired sparsity level.

\subsection{Post-Training Pruning Model with Intra-layer Error Corrections}\label{sec:model}
Post-training compression is typically achieved by decomposing the full-model compression problem into layer-wise subproblems \citep{frantar2023sparsegpt}. 
For instance, a typical Transformer decoder layer \citep{vaswani2017attention} comprises six crucial linear operators: $W_K$, $W_Q$, $W_V$, $W_O$, $W_{fc_1}$, and $W_{fc_2}$. We individually prune each of these operators to eliminate redundant weights while striving to preserve their intended functionality.
Consider a linear operator with weight $W \in \mathbb{R}^{m \times n}$ from the dense model and the corresponding input feature activation $X \in \mathbb{R}^{n \times p}$. 
Its output is computed by $WX \in \mathbb{R}^{m \times p}$. Denoting the pruned counterpart by $W^* \in \mathbb{R}^{m \times n}$, a straightforward approach to quantify the output error is to use the Frobenius norm of the difference between the outputs from the dense and pruned weights
\begin{equation}\label{equ:output_error}
    \left\|W^*X - WX\right\|_F,
\end{equation}
which serves as a metric of the pruning quality at the target sparsity level and is widely adopted by work such as \citep{frantar2023sparsegpt, bovza2024fast}.
\begin{figure}[t]
    \centering
    \includegraphics[width=0.80\linewidth]{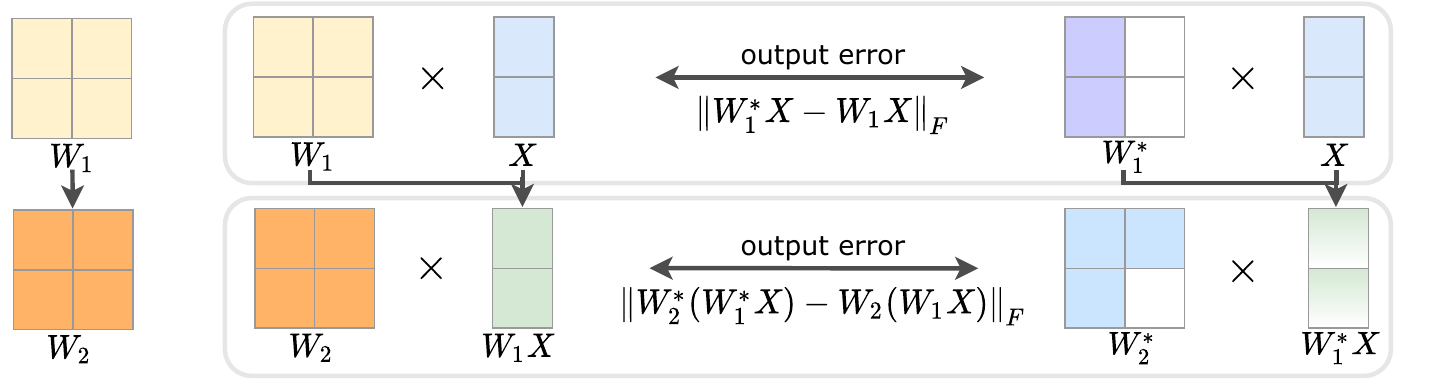}
    \caption{Illustration of the proposed intra-layer error correction mechanism. $W_1$ and $W_2$ represent the weights of two sequential layers within the network architecture.}
    \label{fig:error_correction}
\end{figure}

However, we observe that applying \eqref{equ:output_error} can lead to an error accumulation issue across sequential operators, as illustrated in Figure \ref{fig:error_correction}. In the figure, $W_1$ and $W_2$ represent the weights of two sequential operators. Although \eqref{equ:output_error} effectively quantifies the output error between $W_1$ and its pruned counterpart $W_1^*$ since they are at the top of the layer and share the same inputs, issues arise when applying the same metric to the outputs of $W_2$ and $W_2^*$. Following \eqref{equ:output_error}, the deviation between the outputs of $W_2$ and $W_2^*$ is computed with the same input $W_1X$. However, in a pruned network, the actual input for $W_2^*$ is $W_1^*X$, creating a discrepancy with $W_1X$ and thus leading to error propagation through the operators. To address this, we propose a method to sequentially prune weights within each pruning unit (e.g., a decoder layer of a Transformer), updating \eqref{equ:output_error} to:
\begin{equation}\label{equ:output_error_corre}
\|W^*X^* - WX\|_F,
\end{equation}
where $X^*$ represents the input feature activation for $W^*$ from the sequentially pruned network.

Pruning essentially transforms dense weight matrices into sparse structures. 
The \(\ell_0\)-norm, which directly counts the number of non-zero entries in a vector, is the most straightforward measure of sparsity.
However, because it leads to non-convex and NP-hard optimization problems, we turn to its optimal convex approximation, the \(\ell_1\)-norm \citep{candes2006robust}.
Specifically, to effectively induce sparsity while ensuring computational feasibility in the pruned weights \(W^*\), we apply the \(\ell_1\)-norm to each row of \(W^*\), thereby promoting sparsity throughout the matrix (see Appendix \ref{sec:l1rwo} for detailed explanations):
\begin{equation}\label{equ:l1_norm}
    \left\|W^*_{i, :}\right\|_1, \ i = 1, 2, \dots, m,
\end{equation}
where \(W^*_{i, :}\) represents the \(i\)-th row of \(W^*\). 

Then, we construct our optimization model by integrating \eqref{equ:output_error_corre} and \eqref{equ:l1_norm}
\begin{equation}\label{equ:model}
    \min_{W^* \in \mathbb{R}^{m\times n}} \frac{1}{2} \|W^*X^* - WX\|_F^2 + \lambda \sum_{i=1}^m\|W^*_{i, :}\|_1.
\end{equation}
This model aims to simultaneously minimize both the output error and the sum of the $\ell_1$-norm values while the hyperparameter $\lambda > 0$ balances these two terms. 

\begin{remark}
    The proposed optimization model \eqref{equ:model} is convex. This is due to the fact that the square of the Frobenius norm is a convex function, as is the $\ell_1$-norm. Consequently, the objective function, being a sum of these two convex functions, is also convex. Since the problem is an unconstrained optimization with a convex objective function, the overall optimization model is convex.
\end{remark}

\subsection{Optimization based on FISTA}\label{sec:fista_main_body}
We apply FISTA \citep{beck2009fast} to solve the proposed model \eqref{equ:model} efficiently. Specifically, starting with $t_0 = 1$ and an initial $W^*_0$, the $k$-th iteration of FISTA reads:

\begin{subequations}
\label{eqn:fista_model}
\begin{empheq}[left=\text{FISTA for } \eqref{equ:model}\empheqlbrace]{align}
W^*_{k+\frac{1}{3}} & = W^*_{k} - \frac{1}{L}\left(W^*_{k}X (X^*)^\top - W X (X^*)^\top \right), \label{eqn:step1} \\
W^*_{k+\frac{2}{3}} & = \mathrm{SoftShrinkage}_{\frac{\lambda}{L}}\left(W^*_{k+\frac{1}{3}} \right), \label{eqn:step2} \\
t_{k+1} & = \frac{1}{2} \left(1+\sqrt{1+4t_k^2}\right), \label{eqn:step3} \\
W^*_{k+1} & = W^*_{k+\frac{2}{3}} + \frac{t_k-1}{t_{k+1}}\left( W^*_{k+\frac{2}{3}} - W^*_k \right), \label{eqn:step4}
\end{empheq}
\end{subequations}
where $L = \|X^*(X^*)^\top\|_2$
is the maximum eigenvalue of $X^*(X^*)^\top$ and the $\mathrm{SoftShrinkage}_{\rho}(\cdot)$ operator with parameter $\rho\geq 0$ on a matrix $X=(x_{ij})\in\mathbb{R}^{m\times n}$ performs elementwise transformations defined by
$$
\mathrm{SoftShrinkage}_{\rho}(X) = X', \text{where}\; x'_{ij} =
\begin{cases}
x_{ij} - \rho, \; \text{if} \; x_{ij}>\rho, \\
x_{ij} + \rho, \; \text{if} \; x_{ij}<-\rho, \\
x_{ij} = 0, \; \text{otherwise}.
\end{cases}
$$

Step \eqref{eqn:step1} executes a gradient descent update on the parameter \(W^*_k\), aiming to minimize the function \(1/2\|W^*_kX^* - WX\|_F^2\) with a step size of \(1/L\). Step \eqref{eqn:step2} does a proximal update, defined as:
\begin{equation}
    W^*_{k+\frac{2}{3}}
    =
    \argmin_{W^*}\left\{\frac{L}{2}\left\|W^* - W^*_{k+\frac{1}{3}}\right\|_F^2 + \lambda \sum_{i=1}^m\left\|W^*_{i, :}\right\|_1 \right\}.
\end{equation}

Steps \eqref{eqn:step3} and \eqref{eqn:step4} calculate a linear combination of the previous two points, $\left\{W^*_{k+\frac{2}{3}}, W^*_{k}\right\}$, to facilitate accelerated convergence. Detailed derivations of these steps are provided in Appendix~\ref{sec:fista}.
The FISTA iteration terminates either when the maximum number of iterations, $K$, is reached or when the following stopping criterion is satisfied:
\begin{equation}
\left\|W^*_k - W^*_{k-1}\right\|_F < 1\times 10^{-6}.
\end{equation}
\begin{remark}
    FISTA is proven to achieve a convergence rate of $O(1/k^2)$ \citep{beck2009fast}. This indicates that the distance between the computed solution and the optimal solution decreases proportionally to $1/k^2$ as the number of iterations, $k$, increases.
\end{remark}

Due to the floating-point representation limitations in computers, values near zero computed by FISTA may not be expressed precisely as zero. This imprecision can affect calculations of sparsity, in which exact zeros are counted. To correct this numerical error, a rounding step is implemented after the FISTA iterations to adjust values intended for pruning to exact zeros. Specifically, for the final result $W^*_K$ at the $K$-th iteration and unstructured pruning at sparsity level $s\%$, the rounding step sets the $s\%$ elements with the smallest absolute values in $W^*_K$ to zero. For $n:m$ semi-structured sparsity, the rounding step targets the $n$ elements with the least absolute values within every group of $m$ elements in a row for zeroing. We express the rounding step as:
\begin{equation}\label{equ:rounding}
W^*_{K+1}=\mathrm{round}\left(W^*_K, s\% \text{ or } n:m\right),
\end{equation}
where $\mathrm{round}(\cdot)$ denotes the operation for correcting numerical errors according to the designated sparsity configuration.

\subsection{Adaptive Hyperparameter Tuning}\label{sec:adaptive_algo}
As mentioned, the hyperparameter \(\lambda\) regulates the balance between the two terms in the model \eqref{equ:model}. In addition, the introduction of the rounding step \eqref{equ:rounding} for numerical error corrections also implies that the value of \(\lambda\) indirectly affects the precision loss in this step, especially when the sparsity of \(W^*_K\) achieved through FISTA falls significantly short of the target sparsity. 

Specifically, increasing \(\lambda\) intensifies the focus on the \(\ell_1\)-norm within model \eqref{equ:model}, leading to higher sparsity in \(W^*_K\) but potentially increasing the output error. Conversely, decreasing \(\lambda\) shifts focus towards minimizing the output error, which results in lower sparsity but enhances output accuracy. However, sparsity levels in \(W^*_K\) that are lower than the target can lead to increased rounding errors in~\eqref{equ:rounding}, potentially raising the total error.

To finely adjust \(\lambda\) for minimizing output discrepancy while achieving a target sparsity level, we introduce an adaptive tuning method. We define the total error $\mathcal{E}_{\text{total}}$ and the rounding error $\mathcal{E}_{\text{round}}$ as
\begin{equation}
    \mathcal{E}_{\text{total}} :=\left\|W^*_{K+1}X^* - WX\right\|_F,\; \mathcal{E}_{\text{round}} := \mathcal{E}_{\text{total}} - \left\|W^*_{K}X^* - WX\right\|_F.
\end{equation}

Building on the previous analysis, a high ratio of $\mathcal{E}_{\text{round}} / \mathcal{E}_{\text{total}}$ suggests that a great portion of the error originates from the rounding step \eqref{equ:rounding}, indicating that the sparsity of $W^*_K$ achieved through FISTA is below the target. This implies that the current value of $\lambda$ should be increased to enhance the emphasis on the $\ell_1$-norm in the model \eqref{equ:model}. Conversely, a low ratio of $\mathcal{E}_{\text{round}} / \mathcal{E}_{\text{total}}$ suggests that the achieved sparsity in $W^*_K$ is sufficient, indicating that a reduction in $\lambda$ could shift the focus of model \eqref{equ:model} towards minimizing output error and thus reduce the total error. 

Incorporating the above insights and applying a threshold $\xi$ for $\mathcal{E}_{\text{round}} / \mathcal{E}_{\text{total}}$, we adaptively tune $\lambda$ with the bisection method on $[0, 10^6]$, where $\xi$ is set at 0.3 in our experiments.

\subsection{FISTAPruner Pseudocode}
\renewcommand{\algorithmiccomment}[1]{\hfill{\textcolor{teal}\# #1}}
\begin{algorithm}[t]
\caption{FISTAPruner}
\label{alg:fista_pruner}
\begin{algorithmic}
\STATE \textbf{Inputs:} 
original output $WX$, input activation $X^*$, $\lambda$, $W^*_0$, $K$, $T$, $\epsilon$, $s\% \text{ or } n:m$
\STATE $t \leftarrow 0; \;\; W^*_\text{best} \leftarrow W^*_0; \;\; \mathcal{E}_\text{best} \leftarrow \left\|W^*_{0}X^* - WX\right\|_F$
\REPEAT
\STATE $W^*_K \leftarrow$ $\mathrm{FISTA}\left(WX, X^*, \lambda, W^*_\text{best}, K\right)$ 
\COMMENT{\textcolor{teal}{FISTA iterations as in Section \ref{sec:fista_main_body}}}
\STATE $W^*_{K+1} \leftarrow \mathrm{round}\left(W^*_K, s\% \text{ or } n:m\right)$ \COMMENT{\textcolor{teal}{rounding step for numerical errors as in Section \ref{sec:fista_main_body}}}
\STATE $\mathcal{E}_{\text{total}} \leftarrow \left\|W^*_{K+1}X^* - WX\right\|_F$ \COMMENT{\textcolor{teal}{compute the total error}}
\STATE $\mathcal{E}_{\text{round}} \leftarrow \mathcal{E}_{\text{total}} - \left\|W^*_{K}X^* - WX\right\|_F$ 
\COMMENT{\textcolor{teal}{compute the rounding error}}
\IF{$\mathcal{E}_\text{total} < \mathcal{E}_\text{best}$}
\STATE $W^*_\text{best} \leftarrow W^*_{K+1}$ 
\COMMENT{\textcolor{teal}{preserve the best solution}}
\STATE $\mathcal{E}_\text{stop} = (\mathcal{E}_\text{best}-\mathcal{E}_\text{total})/ \mathcal{E}_\text{best}$ 
\COMMENT{\textcolor{teal}{compute the stop condition}}
\STATE $\mathcal{E}_\text{best} \leftarrow \mathcal{E}_\text{total}$
\COMMENT{\textcolor{teal}{update the best total error}}
\ELSE
\STATE $t \leftarrow t + 1$ 
\COMMENT{\textcolor{teal}{update the number of steps without improvement}}
\ENDIF
\STATE update $\lambda$ by bisection based on $\mathcal{E}_\text{round}/\mathcal{E}_\text{total}$ as in Section \ref{sec:adaptive_algo}
\UNTIL{$t \geq T$ \bf{or} $\mathcal{E}_\text{stop} < \epsilon$}
\STATE \textbf{return} $W^*_\text{best}$
\end{algorithmic}
\end{algorithm}
We treat each decoder layer as an independent pruning unit, enabling parallel pruning across multiple decoder layers on different devices, which significantly enhances the efficiency. Within each decoder layer, the proposed FISTAPruner sequentially prune weights to eliminate error accumulations, as detailed in Section \ref{sec:model}.
Algorithm \ref{alg:fista_pruner} presents FISTAPruner for the dense weight matrix $W$. It leverages FISTA to generate candidate sparse weights based on the model \eqref{equ:model}, as detailed in Section~\ref{sec:fista_main_body}. It then rounds these weights to address numerical errors from floating-point representation and to meet specified sparsity constraints (either $s\%$ unstructured or $n:m$ semi-structured sparsity). Additionally, the parameter $\lambda$ is adaptively tuned, as detailed in Section~\ref{sec:adaptive_algo}, to optimize the trade-off between output error and sparsity. The algorithm iteratively updates the weights, preserving the best solution $W^*_\text{best}$, based on the lowest total error $\mathcal{E}_\text{total}$. It terminates when the number of consecutive iterations without an improvement in $W^*_\text{best}$ reaches $T$, or when the improvement ratio $(\mathcal{E}_\text{best}-\mathcal{E}_\text{total})/ \mathcal{E}_\text{best}$ falls below the threshold $\epsilon$.

\section{Experiments}
In this section, we detail a comprehensive set of experiments designed to validate the efficacy of FISTAPruner. We begin with an in-depth review of our experimental setup. Following this, we explore the perplexity and zero-shot capabilities of the pruned LLMs through rigorous testing and a series of ablation studies. Due to page length constraints, a portion of the results are presented in Appendix \ref{sec:addresultsptb}, \ref{sec:addresultsc4} and \ref{sec:addabs}.

\subsection{Settings}\label{sec:expset}
\textbf{Models.} We utilize models from the OPT \citep{zhang2022opt}, LLaMA \citep{touvron2023llama}, LLaMA-2 \citep{touvron2023llama2}, and LLaMA-3 \citep{meta2023llama3} families. Specifically, we assess our method across OPT-125M/350M/1.3B/2.7B/6.7B/13B/30B, LLaMA-7B/13B/30B/65B, LLaMA-2-7B/13B/70B, and LLaMA-3-8B/70B models.

\textbf{Benchmarks.} Our primary assessment focuses on evaluating the perplexity of pruned LLMs, a metric renowned for its reliability in assessing LLM performance. Following methodologies from previous studies \citep{frantar2023sparsegpt, sun2023simple}, we measure model perplexity using the WikiText-2-raw \citep{merity2016pointer} (hereafter shortened to WikiText), PTB \citep{marcus1994penn}, and C4~\citep{raffel2020exploring} datasets. Additionally, we perform a comprehensive evaluation of the zero-shot capabilities of pruned LLaMA-3-70B models using several standard common-sense benchmark datasets. These include ARC Easy \citep{clark2018think}, ARC Challenge \citep{clark2018think}, WinoGrande \citep{sakaguchi2021winogrande}, BoolQ~\citep{clark2019boolq}, RTE \citep{wang2018glue}, QNLI \citep{wang2018glue}, and WNLI \citep{wang2018glue} tasks, facilitated by the LM Harness library~\citep{gao2021framework}.

\textbf{Baselines.} We compare the FISTAPruner with two state-of-the-art pruning methods as baselines: SparseGPT~\citep{frantar2023sparsegpt} and Wanda~\citep{sun2023simple}. We evaluate two types of sparsity configurations: unstructured and 2:4 semi-structured sparsity.

\textbf{Setup.} We implement FISTAPruner using PyTorch \citep{paszke2019pytorch} and leverage the HuggingFace Transformers library \citep{wolf2019huggingface} for model and dataset management. All pruning experiments are conducted on NVIDIA A100 GPUs, each equipped with 80GB of memory. We observe that FISTAPruner efficiently prunes all LLMs using a single GPU and no more than 40GB of memory. Conversely, SparseGPT can also prune all selected LLMs on a single GPU, while Wanda requires at least four GPUs to prune large models, such as LLaMA-2-70B and LLaMA-3-70B, in their original implementation. For calibration data, we adhere to the approach outlined in previous works \citep{frantar2023sparsegpt, sun2023simple}, utilizing 128 sequences. Each sequence is composed of tokens sampled from the first shard of the C4 dataset, with the number of tokens equal to the maximum embedding length of the LLMs.
For parameters of FISTAPruner, we set the initial value of \(\lambda\) to \(1 \times 10^{-5}\), \(K\) to 20, and \(T\) to 3. For the OPT model family, we use the result of SparseGPT as a warm start for the FISTA iteration and set \(\epsilon\) to \(1 \times 10^{-6}\). For the LLaMA model family, we use the result of Wanda as a warm start and set \(\epsilon\) to \(1 \times 10^{-3}\).

\subsection{Perplexity Experiment Results}
In Tables \ref{tab:ppl_results_opt_wiki} and \ref{tab:ppl_results_llama_wiki}, we present the perplexity results for the pruned OPT, LLaMA, LLaMA-2, and LLaMA-3 models of various sizes on WikiText. For results on PTB and C4, please refer to Appendix~\ref{sec:addresultsptb} and \ref{sec:addresultsc4}. We achieved a 50\% unstructured or 2:4 semi-structured sparsity level by pruning all linear operators, excluding embeddings and the model head. The data in Tables \ref{tab:ppl_results_opt_wiki} and \ref{tab:ppl_results_llama_wiki} illustrate consistent improvements with FISTAPruner over existing methods.
\begin{table*}[ht!]
\centering
\small
\setlength{\tabcolsep}{6.5pt}
\renewcommand{\arraystretch}{1.05}
\resizebox{1.\textwidth}{!}{
\begin{tabular}{l c c c c c c c c}
\toprule 
&  &\multicolumn{7}{c}{OPT} \\
\cmidrule(lr){3-9} 
Method           & Sparsity & 125M           & 350M           & 1.3B           & 2.7B           & 6.7B           & 13B            & 30B  \\
\hline
Dense            & 0$\%$    & 27.66          & 22.00          & 14.63          & 12.47          & 10.86          & 10.13          & 9.56 \\
\hline
SparseGPT        & 50$\%$   & 37.01          & 31.53          & 17.55          & 13.46          & 11.60          & 11.15          & 9.77 \\
Wanda            & 50$\%$   & 38.96          & 36.22          & 18.41          & 14.22          & 11.98          & 11.93          & 10.03 \\
\gr FISTAPruner  & 50$\%$   & \textbf{33.54} & \textbf{28.89} & \textbf{17.21} & \textbf{13.22} & \textbf{11.36} & \textbf{10.95} & \textbf{9.71} \\

\hline
SparseGPT        & 2:4      & 60.02          & 50.15          & 23.83          & 17.20          & 14.13          & 12.94          & 10.92 \\
Wanda            & 2:4      & 80.32          & 113.00         & 28.25          & 21.25          & 15.90          & 15.56          & 13.40 \\
\gr FISTAPruner  & 2:4      & \textbf{45.16} & \textbf{40.41} & \textbf{22.46} & \textbf{15.70} & \textbf{13.16} & \textbf{12.21} & \textbf{10.54} \\
\hline
\end{tabular}
}
\caption{WikiText perplexity ($\downarrow$) of pruned OPT models under 50\% unstructured and 2:4 semi-structured sparsity. FISTAPruner outperforms state-of-the-art methods. 
}
\label{tab:ppl_results_opt_wiki}
\end{table*}

\begin{table*}[ht!]
\centering
\small
\setlength{\tabcolsep}{6.5pt}
\renewcommand{\arraystretch}{1.1}
\resizebox{1.\textwidth}{!}{
\begin{tabular}{l c c c c c c c c c c}
\toprule 
&  &\multicolumn{4}{c}{LLaMA}  & \multicolumn{3}{c}{LLaMA-2} & \multicolumn{2}{c}{LLaMA-3}\\
\cmidrule(lr){3-6} \cmidrule(lr){7-9}\cmidrule(l){10-11}
Method           &Sparsity &7B             &13B            &30B            &65B            & 7B            & 13B           & 70B           & 8B             & 70B \\
\hline
Dense            & 0$\%$   & 5.68          & 5.09          & 4.10          & 3.53          & 5.12          & 4.57          & 3.12          & 5.54           & 2.59 \\
\hline
SparseGPT        & 50$\%$  & 7.24          & 6.22          & 5.33          & 4.60          & 6.54          & 5.63          & 3.99          & 8.64           & 5.30 \\
Wanda            & 50$\%$  & 7.26          & 6.15          & 5.25          & 4.60          & 6.46          & 5.58          & 3.97          & 9.06           & 5.33 \\
\gr FISTAPruner  & 50$\%$  & \textbf{6.97} & \textbf{6.06} & \textbf{5.09} & \textbf{4.39} & \textbf{6.35} & \textbf{5.47} & \textbf{3.93} & \textbf{8.00}  & \textbf{5.09} \\

\hline
SparseGPT        & 2:4     & 11.32         & 9.11          & 7.21          & 6.24          & 10.37         & 8.29          & 5.38          & 14.65          & 8.63 \\
Wanda            & 2:4     & 11.54         & 9.61          & 6.91          & 6.24          & 11.34         & 8.35          & 5.20          & 22.56          & 8.34 \\
\gr FISTAPruner  & 2:4     & \textbf{9.82} & \textbf{8.27} & \textbf{6.70} & \textbf{5.82} & \textbf{9.63} & \textbf{7.69} & \textbf{5.16} & \textbf{14.54} & \textbf{7.55}\\
\hline
\end{tabular}
}
\caption{WikiText perplexity ($\downarrow$) of pruned LLaMA, LLaMA-2 and LLaMA-3 models under 50\% unstructured and 2:4 semi-structured sparsity. FISTAPruner outperforms state-of-the-art methods. 
}
\label{tab:ppl_results_llama_wiki}
\end{table*}

\begin{figure}[t]
    \centering
    \subfigure[Sparsity-vs-perplexity on OPT-125M.]{
        \includegraphics[width=0.45\textwidth]{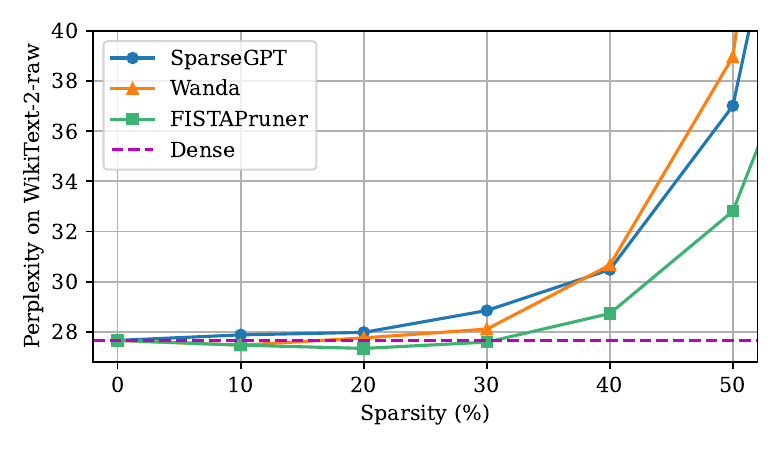}
        \label{fig:125m_10-70}
    }
    \hspace{0.5cm}
    \subfigure[Sparsity-vs-perplexity on LLaMA-3-8B.]{
        \includegraphics[width=0.45\textwidth]{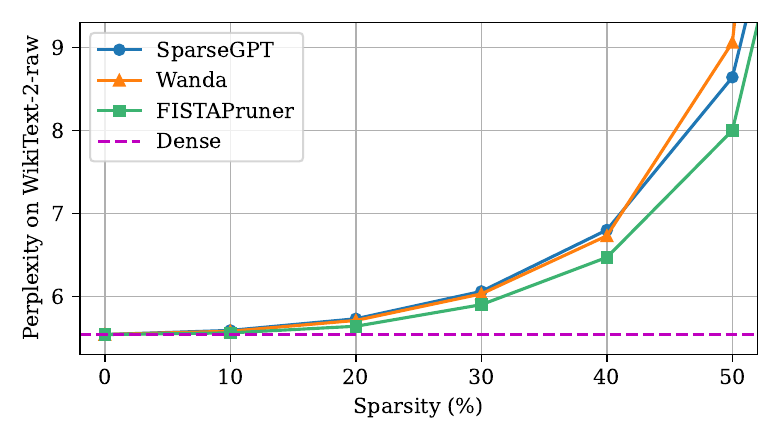}
        \label{fig:8b_10-70}
    }
    \caption{Comparative analysis of sparsity versus perplexity across different methods for OPT-125M and LLaMA-3-8B models on WikiText dataset.}
    \label{fig:10-70}
\end{figure}

To further investigate FISTAPruner's performance under different unstructured sparsity levels, we conducted experiments on the OPT-125M and LLaMA-3-8B models, with perplexity results visualized in Figure \ref{fig:10-70} and measured using WikiText. The results indicate that FISTAPruner consistently outperforms existing methods across different levels of unstructured sparsity. Notably, at 20\% unstructured sparsity on the OPT-125M model, FISTAPruner's performance even surpasses that of the dense network.

\subsection{Zero-Shot Task Results}
The results of zero-shot tasks on pruned LLaMA-3-70B models, with 50\% unstructured and 2:4 semi-structured sparsity, are detailed in Table \ref{tab:zero_shot}. These results indicate that FISTAPruner surpasses existing methods on most tasks. Furthermore, when evaluating the average accuracy across the seven tasks we examined, FISTAPruner consistently shows superior performance compared to existing methods, particularly with 2:4 semi-structured sparsity.

\begin{table*}[ht!]
\centering
\small

\setlength{\tabcolsep}{6.5pt}
\renewcommand{\arraystretch}{1.2}
\resizebox{1.\textwidth}{!}{
\begin{tabular}{l c c c c c c c c c}
\toprule 
Method           & Sparsity & ARC-c           & ARC-e           & WinoGrande      & RTE             & BoolQ           & QNLI            & WNLI            & Mean \\
\hline
Dense            & 0$\%$    & 0.6024          & 0.8685          & 0.8035          & 0.6859          & 0.8560          & 0.5190          & 0.7183          & 0.7219 \\
\hline
SparseGPT        & 50$\%$   & 0.5401          & 0.8340          & 0.7979          & 0.7040          & 0.8480          & 0.5035          & 0.7042          & 0.7045 \\
Wanda            & 50$\%$   & 0.5427          & 0.8320          & 0.7814          & \textbf{0.7076} & 0.8480          & 0.5045          & 0.6338          & 0.6928 \\
\gr FISTAPruner  & 50$\%$   & \textbf{0.5614} & \textbf{0.8410} & \textbf{0.8035} & 0.6895          & \textbf{0.8645} & \textbf{0.5055} & \textbf{0.7183} & \textbf{0.7120} \\

\hline
SparseGPT        & 2:4     & 0.4590          & 0.7830          & 0.7609          & 0.6426          & 0.8165          & 0.4985          & 0.5493          & 0.6443 \\
Wanda            & 2:4     & \textbf{0.4829} & 0.7860          & 0.7174          & 0.6354          & 0.7615          & 0.5390          & 0.6056          & 0.6468 \\
\gr FISTAPruner  & 2:4     & 0.4735          & \textbf{0.7985} & \textbf{0.7751} & \textbf{0.7004} & \textbf{0.8540} & \textbf{0.5675} & \textbf{0.6620} & \textbf{0.6901} \\
\hline
\end{tabular}
}
\caption{Zero-shot results (accuracy, $\uparrow$) of the pruned LLaMA-3-70B model under 50\% unstructured and 2:4 semi-structured sparsity. FISTAPruner outperforms state-of-the-art methods on most of the tasks and yields much higher average accuracies especially under 2:4 semi-structured sparsity.
}
\label{tab:zero_shot}
\end{table*}

\subsection{Ablation Studies}\label{sec:ablation}
We conduct ablation studies to evaluate the intra-layer error correction mechanism of FISTAPruner, the impact of varying the number of calibration samples, and the sensitivity to random seeds. To ensure short iteration times, our experiments are limited to the OPT-125M model and take the result of Wanda as a warm start of FISTA. Additionally, we consistently apply 50\% unstructured sparsity in these studies.

\begin{figure}[t]
    \centering
    \subfigure[Intra-layer error corrections ablation.]{
        \includegraphics[width=0.45\textwidth]{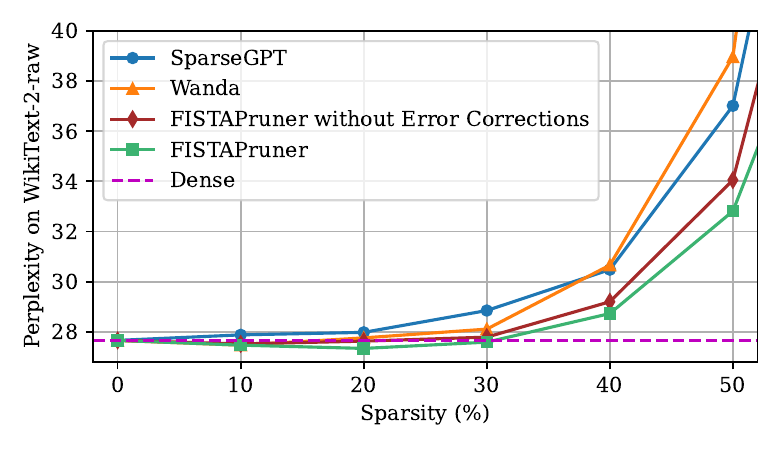}
        \label{fig:error_correction_ablation}
    }
    \hspace{0.5cm}
    \subfigure[Calibration samples ablation.]{
        \includegraphics[width=0.45\textwidth]{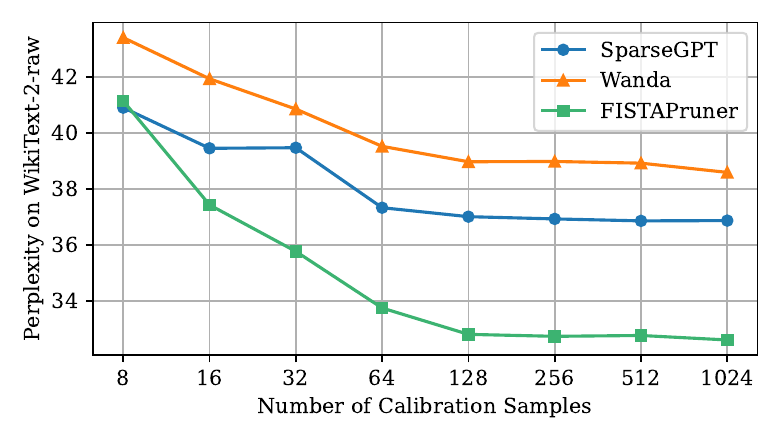}
        \label{fig:nsamples}
    }
    \caption{Ablation studies of FISTAPruner on the WikiText dataset, showcasing the effects of intra-layer error correction and varying calibration sample sizes.}
    \label{fig:ablation_study}
\end{figure}

\textbf{Intra-layer Error Corrections.} We compare the performance of FISTAPruner with and without the intra-layer error correction mechanism, with results on the WikiText dataset displayed in Figure~\ref{fig:error_correction_ablation} (see results on PTB and C4 datasets in Appendix \ref{sec:addabs}). We observe that the perplexity of the pruned model incorporating this mechanism consistently outperforms the version without it, thereby confirming its effectiveness.
Moreover, FISTAPruner, even without the intra-layer error correction mechanism, outperforms existing methods such as SparseGPT and Wanda. This underscores the effectiveness of applying convex optimization theory and algorithms to pruning problems.

\textbf{Amount of Calibration Data.}
We investigate the performance of FISTAPruner and existing methods, SparseGPT and Wanda, in relation to the number of calibration data samples, which we vary in powers of two. The results on WikiText dataset are displayed in Figure \ref{fig:nsamples} (see results on PTB and C4 datasets in Appendix \ref{sec:addabs}).
We observe that using more calibration samples significantly enhances performance, but only up to a certain point as the improvement curve quickly flattens. This finding aligns with observations in \citep{frantar2023sparsegpt, sun2023simple}. Given that using more samples increases computational and memory costs, we consistently use 128 calibration samples in all our experiments.

\textbf{Sensitivity to Random Seeds.}
We assess the sensitivity of FISTAPruner's results to randomness, particularly in relation to the random sampling of calibration data. We conducted five repeated pruning runs using different seeds for data sampling, yielding result of $33.22 \pm 0.361$ (mean $\pm$ standard deviation), which suggests that FISTAPruner demonstrates considerable robustness to variations in the calibration data used.

\section{Discussion}\label{sec:discuss}
Despite the rigorous theoretical foundation and impressive pruning performance of FISTAPruner, the time required for pruning remains a limitation of our method compared to SparseGPT and Wanda. This is primarily due to the iterative nature of FISTA and the process of tuning $\lambda$. Pruning time varies with model size; for instance, it takes about 10 minutes for OPT-125M, while LLaMA-3-70B requires approximately 12 hours on a single Nvidia A100 GPU with 40GB of memory. However, the parallel-pruning capability of FISTAPruner, which allows for simultaneous pruning of multiple decoder layers across various devices, can mitigate this issue to some extent. Furthermore, as post-training pruning is typically an offline process, time sensitivity may not be a critical factor in real-world applications.
In addition, FISTAPruner represents an attempt to integrate convex optimization theory and algorithms into LLM applications, potentially inspiring further advancements in this area.

\section{Conclusion}
In this paper, we introduce FISTAPruner, a layer-wise post-training pruning method for LLMs based on convex optimization models and algorithms. Initially, we develop a convex optimization model that employs the \(\ell_1\)-norm to induce sparsity in the weights, complemented by an intra-layer error correction mechanism to eliminate cumulative errors across operators in the traditional pruning process. Subsequently, we utilize the FISTA solver to efficiently solve the proposed model. FISTAPruner supports both unstructured and \(n:m\) semi-structured pruning and facilitates parallel pruning, which could reduce the total pruning time by utilizing various devices simultaneously. Extensive experiments on the OPT, LLaMA, LLaMA-2, and LLaMA-3 model families demonstrate FISTAPruner’s superior performance compared to existing methods. We hope this work enhances understanding of sparsity in LLMs and inspires further integration of convex optimization within LLM applications.

\newpage
{\small
\bibliography{ref}
\bibliographystyle{apalike}
}

\newpage
\appendix

\section*{Appendix}

\section{Derivations of the Proposed Optimization Model}\label{sec:l1rwo}
We present detailed derivations of Model \eqref{equ:model} in the following.
Given $X^*\in \mathbb{R}^{n\times p}$ and $WX \in \mathbb{R}^{m\times p}$, we want to find a sparse solution $W^*\in \mathbb{R}^{m\times n}$ that minimizes the pruning metric 
\begin{equation}\label{equ:pmetric}
\|W^*X^* - WX\|_F.
\end{equation}
We observe its similarities to the well-known least absolute shrinkage and selection operator (LASSO)~\citep{tibshirani1996regression} problem and thus transform it into a standard LASSO model, which could be efficiently solved by operator-splitting algorithms such as FISTA.
To achieve such a transformation, first, we leverage the following equality to write the decision variable $W^*$ in its vector form:
\begin{align*}
\|W^*X^* - WX\|_F^2 
&= \left\|(X^*)^\top (W^*)^\top -(WX)^\top\right\|_F^2 \\
&= \sum_{i=1}^m \left\| (X^*)^\top (W^*_{i,:})^\top - (WX)_{i,:}^\top \right\|_2^2 \\
&=
\setlength\arraycolsep{0.8pt}
\left\| 
\begin{pmatrix}
(X^*)^\top &      &  \\ 
           & \ddots &  \\ 
           &        & (X^*)^\top \\
\end{pmatrix}
\begin{pmatrix}
(W_{1,:}^*)^\top\\
(W_{2,:}^*)^\top\\
\vdots\\
(W_{m,:}^*)^\top
\end{pmatrix}
-
\begin{pmatrix}
(WX)_{1,:}^\top \\
(WX)_{2,:}^\top \\
\vdots \\
(WX)_{m,:}^\top
\end{pmatrix}
\right\|_2^2
\end{align*}

Then we can rewrite the square of the pruning metric in its vector form,
\begin{equation}\label{equ:model_vec}
\left\| A\mathbf{x} - \mathbf{b} \right\|_2^2,
\end{equation}
where
$$
\setlength\arraycolsep{0.8pt}
A = 
\begin{pmatrix}
(X^*)^\top &      &  \\ 
           & \ddots &  \\ 
           &        & (X^*)^\top \\
\end{pmatrix}
\in \mathbb{R}^{pm\times nm},\;
\mathbf{x} = 
\begin{pmatrix}
(W_{1,:}^*)^\top\\
(W_{2,:}^*)^\top\\
\vdots\\
(W_{m,:}^*)^\top
\end{pmatrix}
\in \mathbb{R}^{nm},\;
\mathbf{b} = 
\begin{pmatrix}
(WX)_{1,:}^\top \\
(WX)_{2,:}^\top \\
\vdots \\
(WX)_{m,:}^\top
\end{pmatrix}
\in \mathbb{R}^{pm}.
$$
Note that finding a sparse $W^*$ to minimize \eqref{equ:pmetric} is equivalent to finding a sparse $\mathbf{x}$ to minimize \eqref{equ:model_vec},
which could be modeled by the LASSO formulation
$$
\min_{\mathbf{x}} \frac12 \left\| A\mathbf{x} - \mathbf{b} \right\|_2^2 + \lambda \|\mathbf{x}\|_1.
$$
Now, we have
\begin{align*}
\frac12 \left\| A\mathbf{x} - \mathbf{b} \right\|_2^2 + \|\mathbf{x}\|_1
&=  \frac12 \|W^*X^* - WX\|_F^2 + 
\lambda 
\left\|
\begin{pmatrix}
(W_{1,:}^*)^\top\\
(W_{2,:}^*)^\top\\
\vdots\\
(W_{m,:}^*)^\top
\end{pmatrix}
\right\|_1\\
&= \frac12 \|W^*X^* - WX\|_F^2 + \lambda \sum_{i=1}^m \left\|(W_{i,:}^*)^\top\right\|_1, 
\end{align*}
and hence, we obtain the proposed optimization model \eqref{equ:model}.

\section{Derivations of the FISTA Iterations}\label{sec:fista}

We derive here the FISTA Iterations for the optimization problem \eqref{equ:model} in which one full iteration includes a gradient descent step of the quadratic term $\frac12 \|W^*X^* - WX\|_F^2$, a proximal step of the regularization term $\lambda \sum_{i=1}^m \left\|(W_{i,:}^*)^\top\right\|_1 $ and a Nestrov acceleration term that yields a improved convergence rate of $O(1/k^2)$~\citep{beck2009fast}. 

Let 
$
f:\mathbb{R}^{m\times n} \rightarrow \mathbb{R}_+ 
$ 
be a function defined by 
$$
f(Y) := \frac12 \left\| YX^* - WX \right\|_F^2.
$$
The gradient of $f$ at $Y = W^*_k$ is computed as
\begin{align*}
\nabla f(W^*_k) 
&= (W^*_kX^* - WX)(X^*)^\top \\
&= W^*_kX^*(X^*)^\top - WX(X^*)^\top.
\end{align*}
Thus, given optimal step size $1/L$ where $L$ is the maximum eigenvalue of $X^*(X^*)^\top$ \citep{beck2009fast}, the gradient descent step \eqref{eqn:step1} of FISTA reads as
$$
W^*_{k+\frac{1}{3}} = W^*_{k} - \frac{1}{L}\left(W^*_{k}X (X^*)^\top - W^\top X (X^*)^\top \right).
$$

In the second step \eqref{eqn:step2}, we do a proximal update with respect to the regularization term by solving 
\begin{equation}\label{equ:proximal}
\min_{W^* \in \mathbb{R}^{m\times n}} 
\frac{L}{2} \left\|W^* - W^*_{k+\frac{1}{3}}\right\|_F^2 + \lambda \sum_{i=1}^m\|W^*_{i, :}\|_1.
\end{equation}
Let 
$
h:\mathbb{R} \rightarrow \mathbb{R}_+ 
$ 
be a function defined by 
$$
h(y|z) := \frac12 (y - z)^2 + \frac \lambda L |y|.
$$
Observe that
$$
\frac{L}{2} \left\|W^* - W^*_{k+\frac{1}{3}}\right\|_F^2 + \lambda \sum_{i=1}^m\|W^*_{i, :}\|_1 
= 
L\sum_{i,j} h\left(W^*_{ij}\Big|W^*_{k+\frac{1}{3}, ij}\right). \\
$$
Hence problem \eqref{equ:proximal} can be split into $m\times n$ independent subproblems of dimension 1 and we only need to focus on solving each one of them.
Note that $h$ is convex but not smooth. It suffices to find a point $W^*_{k+\frac{2}{3}, ij}$ such that 
$$
0 \in \partial h\left(W^*_{k+\frac{2}{3}, ij} \Big |W^*_{k+\frac{1}{3}, ij} \right),
$$
where $\partial$ denotes the sub-differential operator.
Observe that
$$
\partial h(y|z) = 
\begin{cases}
y-z + \frac \lambda L,\; \text{if}\; y > 0, \\
y-z - \frac \lambda L,\; \text{if}\; y < 0, \\
\{y-z+u\frac \lambda L \;|\; u\in [-1,1]\},\; \text{if}\; y=0.
\end{cases}
$$
We now solve for $0\in \partial h(y|z)$ by considering the following cases:
\begin{itemize}
\item If $y>0$, then we set $y-z+\frac \lambda L = 0$. This gives $y = z - \frac\lambda L$ and requires $z > \frac\lambda L$.
\item  If $y<0$, then we set $y-z-\frac \lambda L = 0$. This gives $y = z + \frac\lambda L$ and requires $z < -\frac\lambda L$.
\item If $y=0$, then we want $0 \in \{y-z+u\frac \lambda L \;|\; u\in [-1,1]\}$. This requires $-\frac\lambda L < z < \frac\lambda L$.
\end{itemize}
Hence, $0\in \partial h\left(W^*_{k+\frac{2}{3}, ij} \Big |W^*_{k+\frac{1}{3}, ij} \right)$ yields
$$
W^*_{k+\frac{2}{3}, ij} = 
\begin{cases}
W^*_{k+\frac{1}{3}, ij} - \frac\lambda L, \; \text{if} \; W^*_{k+\frac{1}{3}, ij} > \frac\lambda L, \\
W^*_{k+\frac{1}{3}, ij} + \frac\lambda L, \; \text{if} \; W^*_{k+\frac{1}{3}, ij} < -\frac\lambda L, \\
0, \; \text{otherwise},
\end{cases}
$$
which is exactly the value given by
$
\mathrm{SoftShrinkage}_{\lambda/L}\left(W^*_{k+\frac{1}{3}, ij} \right).
$

Finally, according to \citep{beck2009fast}, we add a Nestrov acceleration step by setting $t_0=1$ and computing
\begin{align}
t_{k+1} & = \frac{1}{2} \left(1+\sqrt{1+4t_k^2}\right), \\
W^*_{k+1} & = W^*_{k+\frac{2}{3}} + \frac{t_k-1}{t_{k+1}}\left( W^*_{k+\frac{2}{3}} - W^*_k \right),
\end{align}
which gives steps \eqref{eqn:step3} and \eqref{eqn:step4}.

The above illustrates the details of the FISTA iterations.

\section{Additional Results}
\subsection{Perplexity Results on PTB}\label{sec:addresultsptb}

We present the PTB perplexity results of pruned OPT, LLaMA, LLaMA-2, and LLaMA-3 models under 50\% unstructured and 2:4 semi-structured sparsity in Tables \ref{tab:ppl_results_opt_ptb} and \ref{tab:ppl_results_llama_ptb}.
FISTAPruner outperforms state-of-the-art methods on all OPT, LLaMA and LLaMA-3 models, as well as on most LLaMA-2 models on the PTB dataset. 
The sole exception is the pruning of the LLaMA-2-70B model under 50\% unstructured sparsity, where FISTAPruner surpasses Wanda but falls short of SparseGPT. 
This underperformance may be due to the generally poorer performance of LLaMA-2 models compared to similarly sized models from other families. 
For instance, the dense LLaMA-2-13B model exhibits a PTB perplexity of 56.52, even higher than the smaller LLaMA-2-7B model, which has a perplexity of 50.19. 
Moreover, we observe that the PTB perplexity results for all dense LLaMA and LLaMA-2 models are consistently higher than those for similarly sized OPT models; for example, the LLaMA-2-13B's perplexity of 56.52 far exceeds the smallest OPT-125M model's 38.99. 
In contrast, LLaMA-3 models show significantly better performance on the PTB dataset.

\begin{table*}[ht!]
\centering
\small
\setlength{\tabcolsep}{6.5pt}
\renewcommand{\arraystretch}{1.0}
\resizebox{1.\textwidth}{!}{
\begin{tabular}{l c c c c c c c c}
\toprule 
&  &\multicolumn{7}{c}{OPT} \\
\cmidrule(lr){3-9} 
Method           & Sparsity & 125M            & 350M           & 1.3B           & 2.7B           & 6.7B           & 13B            & 30B  \\
\hline
Dense            & 0$\%$    & 38.99           & 31.07          & 20.29          & 17.97          & 15.77          & 14.52          & 14.04 \\
\hline
SparseGPT        & 50$\%$   & 55.38           & 43.58          & 25.64          & 20.52          & 17.38          & 15.98          & 14.97 \\
Wanda            & 50$\%$   & 57.60           & 55.47          & 27.98          & 21.85          & 17.92          & 17.45          & 15.47 \\
\gr FISTAPruner  & 50$\%$   & \textbf{49.79}  & \textbf{41.26} & \textbf{25.08} & \textbf{20.15} & \textbf{17.08} & \textbf{15.87} & \textbf{14.92} \\

\hline
SparseGPT        & 2:4      & 94.21           & 72.82          & 37.30          & 26.87          & 21.65          & 18.69          & 16.56 \\
Wanda            & 2:4      & 111.55          & 135.98         & 43.85          & 34.64          & 25.07          & 22.16          & 21.65 \\
\gr FISTAPruner  & 2:4      & \textbf{67.80}  & \textbf{59.51} & \textbf{36.26} & \textbf{24.43} & \textbf{20.04} & \textbf{18.08} & \textbf{16.18} \\
\hline
\end{tabular}
}
\caption{PTB perplexity  of pruned OPT models under 50\% unstructured and 2:4 semi-structured sparsity. FISTAPruner outperforms state-of-the-art methods. }
\label{tab:ppl_results_opt_ptb}
\end{table*}

\begin{table*}[ht!]
\centering
\small
\setlength{\tabcolsep}{6.5pt}
\renewcommand{\arraystretch}{1.05}
\resizebox{1.\textwidth}{!}{
\begin{tabular}{l c c c c c c c c c c}
\toprule 
&  &\multicolumn{4}{c}{LLaMA}  & \multicolumn{3}{c}{LLaMA-2} & \multicolumn{2}{c}{LLaMA-3}\\
\cmidrule(lr){3-6} \cmidrule(lr){7-9}\cmidrule(l){10-11}
Method           &Sparsity & 7B             & 13B            & 30B            & 65B            & 7B              & 13B            & 70B            & 8B              & 70B \\
\hline
Dense            & 0$\%$   & 41.15          & 28.10          & 23.51          & 25.07          & 50.19           & 56.52          & 22.68          & 10.17           & 7.87 \\
\hline
SparseGPT        & 50$\%$  & 79.67          & 37.49          & 26.14          & 27.64          & 1020.01             & 95.41          & \textbf{24.87} & 14.00           & 9.24 \\
Wanda            & 50$\%$  & 80.48          & 36.43          & 26.64          & 25.77          & 97.58           & 86.79          & 26.07          & 15.54           & 9.44 \\
\gr FISTAPruner  & 50$\%$  & \textbf{58.67} & \textbf{35.30} & \textbf{25.63} & \textbf{25.15} & \textbf{96.72}  & \textbf{78.23} & 25.36          & \textbf{12.93}  & \textbf{8.88} \\

\hline
SparseGPT        & 2:4     & 154.62         & 71.68          & 32.44          & 32.91          & 1163.57             & 154.15          & 31.51          & 23.42          & 13.01 \\
Wanda            & 2:4     & 211.40         & 74.29          & 35.56          & 33.39          & 587.54          & 224.55          & 33.97          & 48.96          & 14.17 \\
\gr FISTAPruner  & 2:4     & \textbf{91.84} & \textbf{64.04} & \textbf{30.86} & \textbf{30.78} & \textbf{361.16} & \textbf{136.84} & \textbf{31.49} & \textbf{22.60} & \textbf{11.11}\\
\hline
\end{tabular}
}
\caption{PTB perplexity ($\downarrow$) of pruned LLaMA, LLaMA-2 and LLaMA-3 models under 50\% unstructured and 2:4 semi-structured sparsity. }
\label{tab:ppl_results_llama_ptb}
\end{table*}

\subsection{Perplexity Results on C4}\label{sec:addresultsc4}

The C4 perplexity results of pruned OPT, LLaMA, LLaMA-2, and LLaMA-3 models under 50\% unstructured and 2:4 semi-structured sparsity are shown in Tables \ref{tab:ppl_results_opt_ptb} and \ref{tab:ppl_results_llama_ptb}.
FISTAPruner performs consistently better than the state-of-the-art methods.

\begin{table*}[ht]
\centering
\small
\setlength{\tabcolsep}{6.5pt}
\renewcommand{\arraystretch}{1.05}
\resizebox{1.\textwidth}{!}{
\begin{tabular}{l c c c c c c c c}
\toprule 
&  &\multicolumn{7}{c}{OPT} \\
\cmidrule(lr){3-9} 
Method           & Sparsity & 125M           & 350M           & 1.3B           & 2.7B           & 6.7B           & 13B            & 30B  \\
\hline
Dense            & 0$\%$    & 26.56          & 22.59          & 16.07          & 14.34          & 12.71          & 12.06          & 11.45 \\
\hline
SparseGPT        & 50$\%$   & 33.52          & 29.14          & 19.23          & 15.77          & 13.73          & 12.98          & 11.96 \\
Wanda            & 50$\%$   & 34.89          & 34.46          & 20.63          & 16.44          & 14.25          & 13.57          & 12.32 \\
\gr FISTAPruner  & 50$\%$   & \textbf{30.93} & \textbf{27.36} & \textbf{18.56} & \textbf{15.58} & \textbf{13.61} & \textbf{12.94} & \textbf{11.92} \\

\hline
SparseGPT        & 2:4      & 52.11          & 46.36         & 25.77          & 19.35          & 16.44          & 14.85          & 13.18 \\
Wanda            & 2:4      & 64.73          & 88.62         & 28.59          & 22.88          & 19.00          & 16.19          & 16.18 \\
\gr FISTAPruner  & 2:4      & \textbf{38.08} & \textbf{36.45} & \textbf{24.29} & \textbf{17.82} & \textbf{15.35} & \textbf{14.19} & \textbf{12.78} \\
\hline
\end{tabular}
}
\caption{C4 perplexity ($\downarrow$) of pruned OPT models under 50\% unstructured and 2:4 semi-structured sparsity. FISTAPruner outperforms state-of-the-art methods. 
}
\label{tab:ppl_results_opt_c4}
\end{table*}

\begin{table*}[ht!]
\centering
\small
\setlength{\tabcolsep}{6.5pt}
\renewcommand{\arraystretch}{1.1}
\resizebox{1.\textwidth}{!}{
\begin{tabular}{l c c c c c c c c c c}
\toprule 
&  &\multicolumn{4}{c}{LLaMA}  & \multicolumn{3}{c}{LLaMA-2} & \multicolumn{2}{c}{LLaMA-3}\\
\cmidrule(lr){3-6} \cmidrule(lr){7-9}\cmidrule(l){10-11}
Method           &Sparsity & 7B             & 13B           & 30B           & 65B           & 7B             & 13B            & 70B           & 8B              & 70B \\
\hline
Dense            & 0$\%$   & 7.34           & 6.80          & 6.13          & 5.81          & 7.04           & 6.52           & 5.53          & 9.01            & 6.82 \\
\hline
SparseGPT        & 50$\%$  & 9.33           & 8.14          & 7.34          & 6.66          & 9.00           & 7.96           & 6.25          & 13.93           & 9.34 \\
Wanda            & 50$\%$  & 9.34           & 8.15          & 7.29          & 6.71          & 8.94           & 8.04           & 6.30          & 14.97           & 9.80 \\
\gr FISTAPruner  & 50$\%$  & \textbf{8.90}  & \textbf{7.96} & \textbf{7.05} & \textbf{6.49} & \textbf{8.62}  & \textbf{7.73}  & \textbf{6.22} & \textbf{13.12}  & \textbf{8.94} \\

\hline
SparseGPT        & 2:4     & 13.65          & 11.38          & 9.50          & 8.41         & 13.58          & 11.39          & 7.99          & 24.16          & 14.81 \\
Wanda            & 2:4     & 14.47          & 12.11          & 9.46          & 8.78          & 15.07          & 12.13          & 7.89          & 36.70          & 14.47 \\
\gr FISTAPruner  & 2:4     & \textbf{11.95} & \textbf{10.27} & \textbf{8.81} & \textbf{7.82} & \textbf{12.41} & \textbf{10.34} & \textbf{7.59} & \textbf{23.15} & \textbf{12.18}\\
\hline
\end{tabular}
}
\caption{C4 perplexity ($\downarrow$) of pruned LLaMA, LLaMA-2 and LLaMA-3 models under 50\% unstructured and 2:4 semi-structured sparsity. FISTAPruner outperforms state-of-the-art methods. 
}
\label{tab:ppl_results_llama_c4}
\end{table*}

\subsection{Additional Ablation Study Results}\label{sec:addabs}
\begin{figure}[ht]
    \centering
    \subfigure[Intra-layer error corrections ablation.]{
        \includegraphics[width=0.45\textwidth]{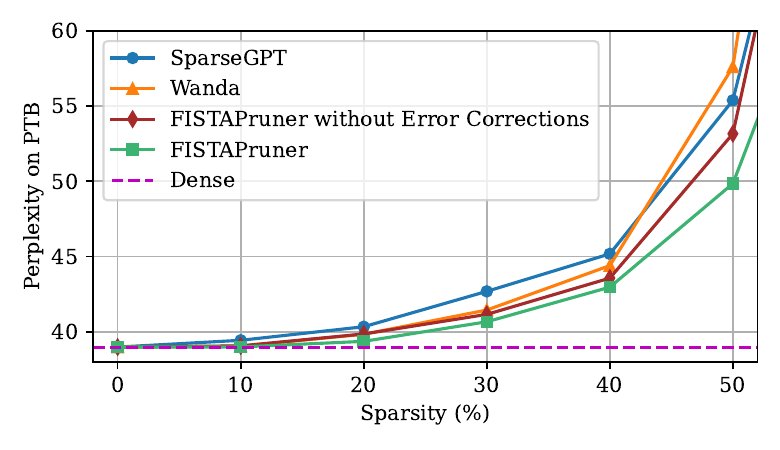}
        \label{fig:error_correction_ablation_ptb}
    }
    \hspace{0.5cm}
    \subfigure[Calibration samples ablation.]{
        \includegraphics[width=0.45\textwidth]{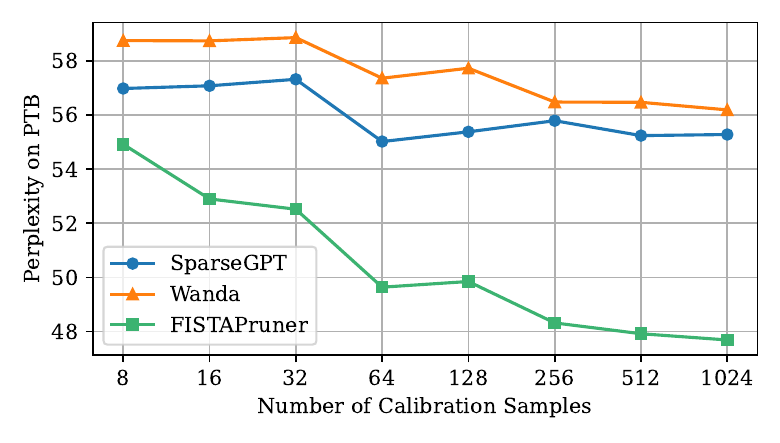}
        \label{fig:nsamples_ptb}
    }
    \caption{Ablation studies of FISTAPruner on the PTB dataset, showcasing the effects of intra-layer error correction and varying calibration sample sizes.}
    \label{fig:ablation_study_ptb}
\end{figure}

\begin{figure}[ht]
    \centering
    \subfigure[Intra-layer error corrections ablation.]{
        \includegraphics[width=0.45\textwidth]{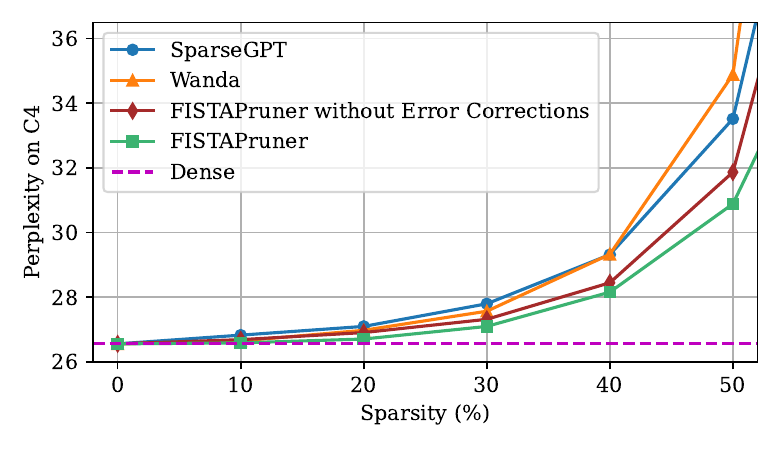}
        \label{fig:error_correction_ablation_c4}
    }
    \hspace{0.5cm}
    \subfigure[Calibration samples ablation.]{
        \includegraphics[width=0.45\textwidth]{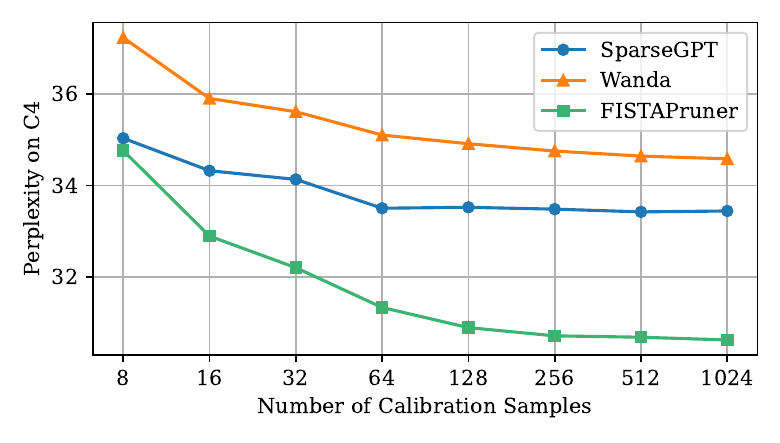}
        \label{fig:nsamples_c4}
    }
    \caption{Ablation studies of FISTAPruner on the C4 dataset, showcasing the effects of intra-layer error correction and varying calibration sample sizes.}
    \label{fig:ablation_study_c4}
\end{figure}
\textbf{Intra-layer Error Corrections.}
We compare the performance of FISTAPruner with and without the intra-layer error correction mechanism, with results on PTB and C4 datasets displayed in Figures \ref{fig:error_correction_ablation_ptb} and \ref{fig:error_correction_ablation_c4}. The results indicate that the perplexity of the pruned model incorporating this mechanism consistently outperforms the version without it, thereby confirming its effectiveness.

\textbf{Amount of Calibration Data.}
The results of pruning performance in relation to the number of calibration data samples on PTB and C4 datasets are displayed in Figures \ref{fig:nsamples_ptb} and \ref{fig:nsamples_c4}. The same curve pattern as shown in Figure \ref{fig:nsamples} is observed.

\end{document}